\documentclass[11pt,fleqn]{scrartcl}

\usepackage[margin=2.0cm]{geometry}

\usepackage[utf8]{inputenc}
\usepackage[T1]{fontenc}
\usepackage[english]{babel}

\usepackage{graphicx}
\usepackage{subcaption}
\usepackage{authblk}
\usepackage[justification=justified]{caption}
\usepackage{gensymb}
\usepackage{siunitx}
\usepackage[version=3]{mhchem}

\usepackage{amsfonts,amsmath,amssymb}
\usepackage{txfonts}

\usepackage{booktabs}
\usepackage{multirow}
\usepackage{listings}
\usepackage{glossaries}
\usepackage{tabularx}
\usepackage{float}
\usepackage{balance}
\usepackage{color,soul}
\usepackage[htt]{hyphenat}
\usepackage[normalem]{ulem}

\usepackage{ragged2e}
\usepackage{xurl}

\usepackage{hyperref}
\usepackage{xr-hyper}

\usepackage[square,numbers,sort&compress]{natbib}
\newcommand{\figref}[1]{Fig.~\ref{#1}}

\newcommand{\keywords}[1]{%
  \vspace{0.5em}%
  \noindent\textbf{Keywords: }#1%
}

\title{GENIUS: An Agentic AI Framework for Autonomous Design and Execution of Simulation Protocols}

\author[1]{Mohammad Soleymanibrojeni}
\author[2]{Roland Aydin}
\author[3]{Diego Guedes-Sobrinho}
\author[4]{Alexandre C.~Dias}
\author[5]{Maur\'icio J.~Piotrowski}
\author[1]{Wolfgang Wenzel}
\author[1,*]{Celso Ricardo Caldeira R\^ego}

\affil[1]{Karlsruhe Institute of Technology, 
          Institute of Nanotechnology Hermann-von-Helmholtz-Platz, 
          Karlsruhe, 76021, Germany}

\affil[2]{Hamburg University of Technology, 
          Eißendorfer Straße 42, Hamburg, 21073, Germany}

\affil[3]{Federal University of Paraná, 
          Department of Chemistry, 
          Curitiba, 
          81531-980, 
          Brazil}

\affil[4]{University of Brasília, 
          Institute of Physics and International Center of Physics, 
          Brasília, 
          70919-970, 
          Brazil}

\affil[5]{Federal University of Pelotas, 
          Department of Physics, 
          Pelotas, 
          96010-900, 
          Brazil}

\affil[*]{\textit{Corresponding author}: \texttt{celso.rego@kit.edu}}

\date{} 

\begin{document}

\maketitle


\begin{abstract}
Predictive atomistic simulations have propelled materials discovery, yet routine setup and debugging still demand computer specialists. This know-how gap limits Integrated Computational Materials Engineering (ICME), where state-of-the-art codes exist but remain cumbersome for non-experts. We address this bottleneck with GENIUS, an AI-agentic workflow that fuses a smart Quantum ESPRESSO knowledge graph with a tiered hierarchy of large language models supervised by a finite-state error-recovery machine. Here we show that GENIUS translates free-form human-generated prompts into validated input files that run to completion on $\approx \SI{80}{\percent}$ of \num{295} diverse benchmarks, where \SI{76}{\percent} are autonomously repaired, with success decaying exponentially to a \SI{7}{\percent} baseline. Compared with LLM-only baselines, GENIUS halves inference costs and virtually eliminates hallucinations. The framework democratizes electronic-structure DFT simulations by intelligently automating protocol generation, validation, and repair, opening large-scale screening and accelerating ICME design loops across academia and industry worldwide.
\end{abstract}


\section*{Introduction}
Computational simulations have revolutionized materials design, accelerating innovation by allowing researchers to explore material properties and their behaviors virtually before experimental validation\cite{hafner2006toward, shen2023computational, bock2019review, Rego_2022}. This shift has led to significant breakthroughs that range from energy storage\cite{Yu_2025, Lin_2024} to pharmaceutical development\cite{Han_2024, King_2025}. However, a persistent challenge undermines this potential: the technical barriers to effective simulation setup disproportionately burden researchers, particularly those whose expertise lies in experimental rather than computational domains. When scientists identify a promising new compound, understanding its fundamental properties often requires computational validation. Yet, even seemingly straightforward simulations frequently lead to lengthy technical challenges. Even experienced computational scientists (physicists, chemists, engineers) find themselves diverted from scientific inquiry toward navigating complex programming challenges, engaging in trial-and-error attempts, and struggling with computational setup details rather than focusing on the scientific questions\cite{Schaarschmidt2021}.

Integrated Computational Materials Engineering (ICME) has emerged as a robust framework to accelerate materials development by synergizing experimental data, simulations, and theoretical models across multiple scales. ICME aims to streamline the transition from laboratory discovery to industrial application through predictive modeling\cite{allison2006integrated,taylor2018integrated}. However, its full potential remains constrained by what implementation science describes as \emph{know-do gap}, the disparity between available computational tools and their practical application by the broader scientific community. This gap persists despite the availability of a wide range of these tools. The open source community has made remarkable progress in the accuracy and consistency of computational codes. Recent community-wide benchmarks demonstrate that modern DFT codes and pseudopotentials now yield near-identical equations of state for elemental crystals, approaching experimental precision\cite{Lejaeghere_2016, Huber_2021}. This technical convergence suggests that the tools are mature, yet the human interface to these tools remains a significant bottleneck. The time spent on technical implementation rather than scientific thinking dramatically slows the pace of discovery. 

As current approaches to the simulation protocol creation rely on predefined or rigid parameter settings across several programs\cite{deAraujo2024, Bonacci_2023, Soleymanibrojeni2024}, the complexity of integrating different computational tools remains challenging. Creating and validating these protocols requires that the user manually interact with the databases to collect relevant data. Furthermore, users must technically master several computational tools' syntax and possess deep expertise in all of them, practically becoming experts in the documentation by debugging protocols when errors occur. Although state-of-the-art (SOTA) electronic-structure codes are accurate, open, and widely accessible, their routine use still demands technical expertise that many domain scientists find challenging. This expertise barrier narrows the pool of researchers who can exploit computational materials science. It creates barriers that drain time and delay critical discoveries, slowing theory translation into concrete advances in batteries, catalysts, and structural alloys. Implementation science principles\cite{luke2024bridges,toner2024artificial}, the field that studies how evidence-based practices move from the laboratory into everyday use, offer a path forward. By diagnosing and dismantling the educational, cultural, and infrastructural obstacles that restrict the adoption of evidence-based tools, this principle can shrink the \emph{know–do gap} separating mature computational capabilities from the needs of working material scientists. The goal is not to invent another method but to democratize the already powerful methods, freeing researchers to focus on scientific questions rather than software configuration and workflow maintenance. In doing so, we can accelerate how computational insights can become real-world materials innovations, which might otherwise transform industries and address global challenges.

\begin{figure}[h!]
    \centering
    \resizebox{0.99\textwidth}{!}{%
        \includegraphics{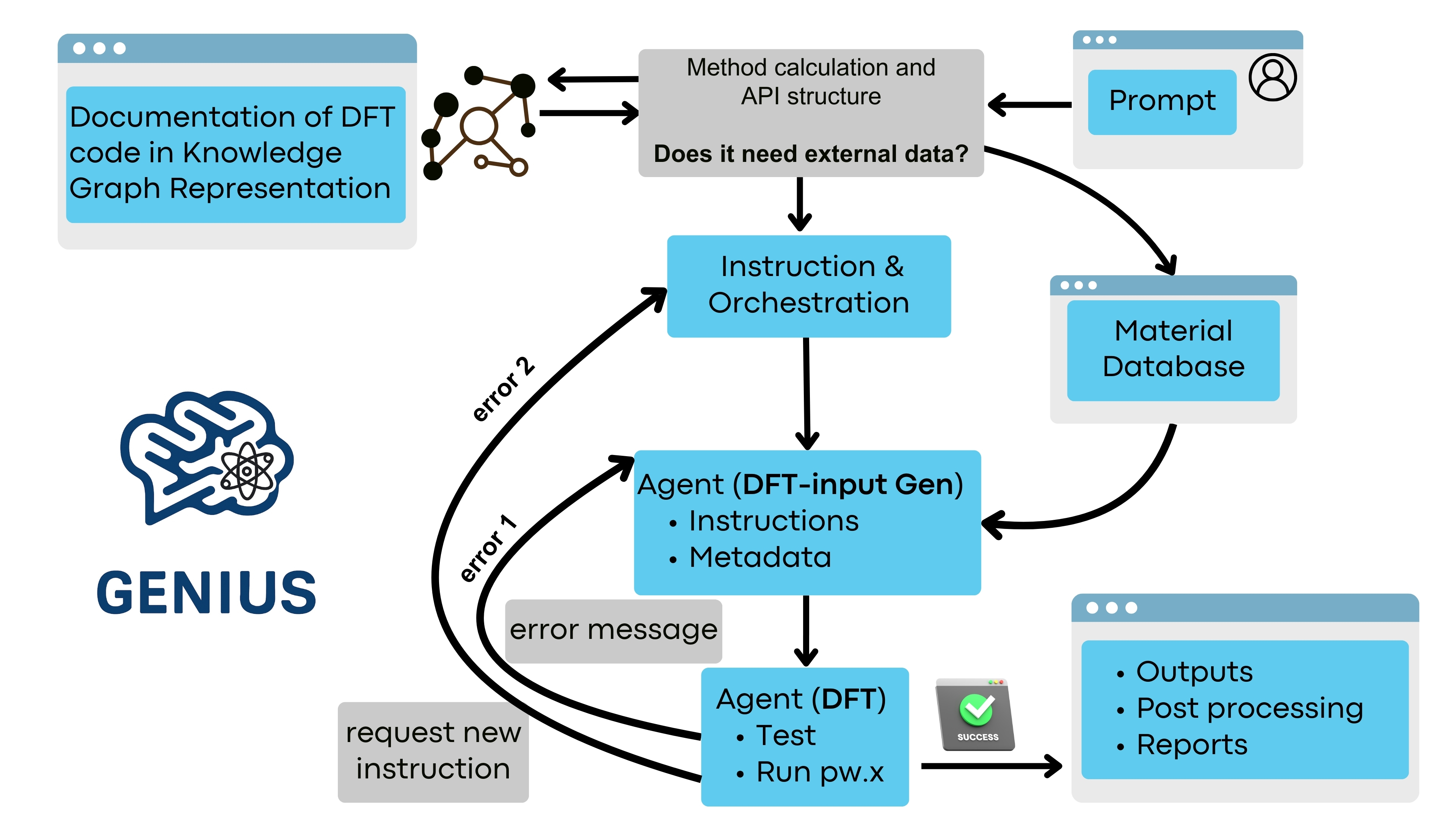}
    }
    \caption{ 
GENIUS framework for autonomous Quantum ESPRESSO simulations. This schematic illustration depicts the end-to-end workflow of the GENIUS framework, designed to overcome technical barriers in DFT simulations. Users' Natural language prompts are interpreted by a recommendation system powered by a smart knowledge graph that encodes Quantum ESPRESSO parameter details and constraints. Large language models then generate the corresponding simulation protocols. The framework includes automated validation and an Automated Error Handling (\texttt{AEH} (\textbf{error 1} and \textbf{error 2}) loop that utilizes a knowledge graph and Large language models to diagnose and correct failed runs iteratively. This integrated process autonomously translates user intent into validated Quantum ESPRESSO input files, ready for submission to the available computational resources.}
    \label{fig:overview}
\end{figure}

To address this critical interface bottleneck and bridge the \emph{know-do gap}, this paper introduces GENIUS, an AI-driven agentic framework combining large language models (LLMs)\cite{jablonka202314} with a smart knowledge graph (KG)\cite{wang2017knowledge} employing reinforcement learning with verifiable rewards\cite{Lambert_2024, DeepSeek-AI_2025, team_2025}, that in our case interprets the schema, enforces constraints, reformulates questions, and surfaces only consistent answers. This framework acts as an intelligent interface to computational tools, specifically designed here for generating and automatically debugging simulation protocols based on Density Functional Theory (DFT), and to validate our approach, we choose the Quantum ESPRESSO (QE) program\cite{Giannozzi_2009}. A schematic overview of this framework is presented in \figref{fig:overview}. The mechanisms of the smart KG with the LLMs infer relevant parameters by understanding both explicit and implicit conditions within the user's request, then critically evaluate the retrieved information to ensure its suitability and context. This process provides accurate and structured knowledge to the LLM, mitigating limitations such as hallucinations \cite{michelutti2024systematic,ledger2024detecting} and enabling reliable protocol generation tailored to user requirements, including material specifics from curated databases. Furthermore, the framework incorporates robust automated error handling capable of debugging and validating protocols when initial attempts fail, overcoming a significant hurdle during the generation of the protocols in practical simulation workflows. This integrated approach addresses the inherent limitations of current AI models when applied to precise scientific tasks.

GENIUS reshapes who can participate in computational materials research by bridging technical computational tasks with the nature of the materials. We connect previously discrete manual tasks into one integrated AI system by evaluating the researcher's request, generating compliant simulation protocols, validating, and handling errors automatically. This integration improves the reproducibility, re-usability, and transferability of simulation protocols\cite{Schaarschmidt2021, gundersen2021fundamental} while making advanced simulations accessible to researchers regardless of their computational background. Our work advances both in ICME and in the implementation of science objectives: we enhance the ICME paradigm by making its computational tools more accessible, while applying implementation science principles to overcome adoption barriers in computational materials research. By allowing researchers to focus on scientific questions rather than technical implementation, GENIUS delivers incremental efficiency gains while enabling a fundamental shift in how—and by whom—materials discovery can be conducted\cite{holbrook2019open}. In the following sections, we detail the creation of the smart QE KG and explain the framework's architecture, including its specialized agents and subsystems, such as recommendation, protocol generation, and automated error handling. We present benchmarks assessing the framework's performance across several computational tasks.

\section*{Methods}
We developed GENIUS to overcome the technical challenges in computational materials science and made the QE code simulation methods more accessible to researchers within the domain-knowledge context. Although we focus here on QE as a representative case, this approach can easily be extended to any atomistic simulation code, including molecular dynamics. This framework combines LLMs with a smart KG, serving as an intelligent interface between users and QE to automate the generation, validation, and debugging of complex simulation protocols. Our system architecture comprises three main components, which are: ($i$) a recommendation system, ($ii$) a protocol generation module, and ($iii$) an automated error handling system (denoted as \textbf{error 1} and \textbf{error 2}) as shown in \figref{fig:overview}. These components are designed to handle specific aspects of the workflow, from user input to final protocol generation. The framework prioritizes reproducibility, accuracy, and user adaptability, aligning with computational materials research's best practices and scientific standards. Therefore, references to computational resources, model names, service providers, or other entities are made by name, as these are considered common knowledge within the relevant fields.

\subsection*{System Architecture Overview}
In GENIUS, our three-tier QE simulation protocol generation architecture is based on a set of sequential propositions. ($i$) Recommendation System: This is the interface between the user and the protocol generation pipeline. It comprises submodules responsible for collecting material-specific information, retrieving relevant QE simulation parameters, and evaluating them. This information is formulated into a template recommended to the protocol generation system. The template includes a collection of parameters with suggested values and data types (\texttt{CHARACTER}, \texttt{REAL}, \texttt{INTEGER}, \texttt{LOGICAL}), followed by materials-specific information like atomic positions and appropriate element-specific pseudopotentials. ($ii$) Protocol Generation System: It generates the simulation protocol based on the template provided by the recommendation system. The system may not utilize all suggested parameters, as the final selection occurs during protocol generation, considering the user's prompt and additional context provided to the LLM. The generated input is syntactically validated by running the QE program. Upon successful validation, the workflow concludes. If validation fails, the automated error handling module is triggered. ($iii$) Automated Error Handling System: It receives the current generated simulation protocol and the error messages from the QE program, and extracts relevant documentation based on these messages. Using the currently generated protocol and extracted documentation, the error can be resolved, and the protocol can be revalidated. Each LLM is allocated a specific number of attempts to resolve the error. If the error persists, the system switches to the next, presumably more capable LLM in a predefined hierarchy. If all attempts fail, the process terminates with a failure message.

\subsubsection*{System Components Integration}
The framework employs a finite state machine (FSM) architecture to manage component interactions and the workflow progression. Each component operates as a distinct FSM state node, with clearly defined transition conditions depending on the state of the executed process. The FSM implementation has an ($i$) Entry Node: to initialize the framework and validate user inputs, ($ii$) State Transitions: defined by the success or failure conditions at each processing step, ($iii$) Error Recovery States: implementing retry mechanisms (for I/O, network, validation), and ($iV$) Terminal States: indicating either Success (valid protocol generation) or Failure (unresolved error). The overall framework is illustrated in \figref{fig:wf-overview}.
\begin{figure}[h!]
    \centering
    \resizebox{0.99\textwidth}{!}{%
        \includegraphics{figures/wf.png}
    }
    \caption{State diagram of the AI-driven framework for generating QE simulation protocols. The diagram is organized into four composite states (dashed boxes): RecommendationSystem parses the user's natural-language request (`Interface' → `InitializeWorkflow'), retrieves materials data (`MaterialsDb') and simulation parameters (via `DocumentCollection' → `ConditionExtraction' → `RetrieveCandidateParameters'), and evaluates them (`EvaluateParameters') to produce a structured input template. `ProtocolGeneration' uses that template (`PrepareInputTemplate') to generate the actual QE input file (`QeInputGeneration'). `ExecutionValidation' runs the simulation (`QeRun'), transitioning to `Finished' on success. `AutomatedErrorHandling' detects failures (`FailureDetected' → `CheckRetries') and either retries execution (`AttemptCorrection'), switches to an alternative model (`SwitchModel'), or terminates at `Failure' if all options are exhausted. Solid arrows show transitions labeled with the triggering action or condition; color and class styling differentiate data sources, main processes, and error-handling loops; more details can be found in the GitHub \href{https://github.com/KIT-Workflows/agentic-workflow-framework}{repository}.
}
    \label{fig:wf-overview}
\end{figure}

\subsection*{Smart Knowledge Graph}

\begin{figure}[h!]
    \centering
        \begin{subfigure}[c]{0.98\textwidth}
        \centering
        \includegraphics[ width=\textwidth]{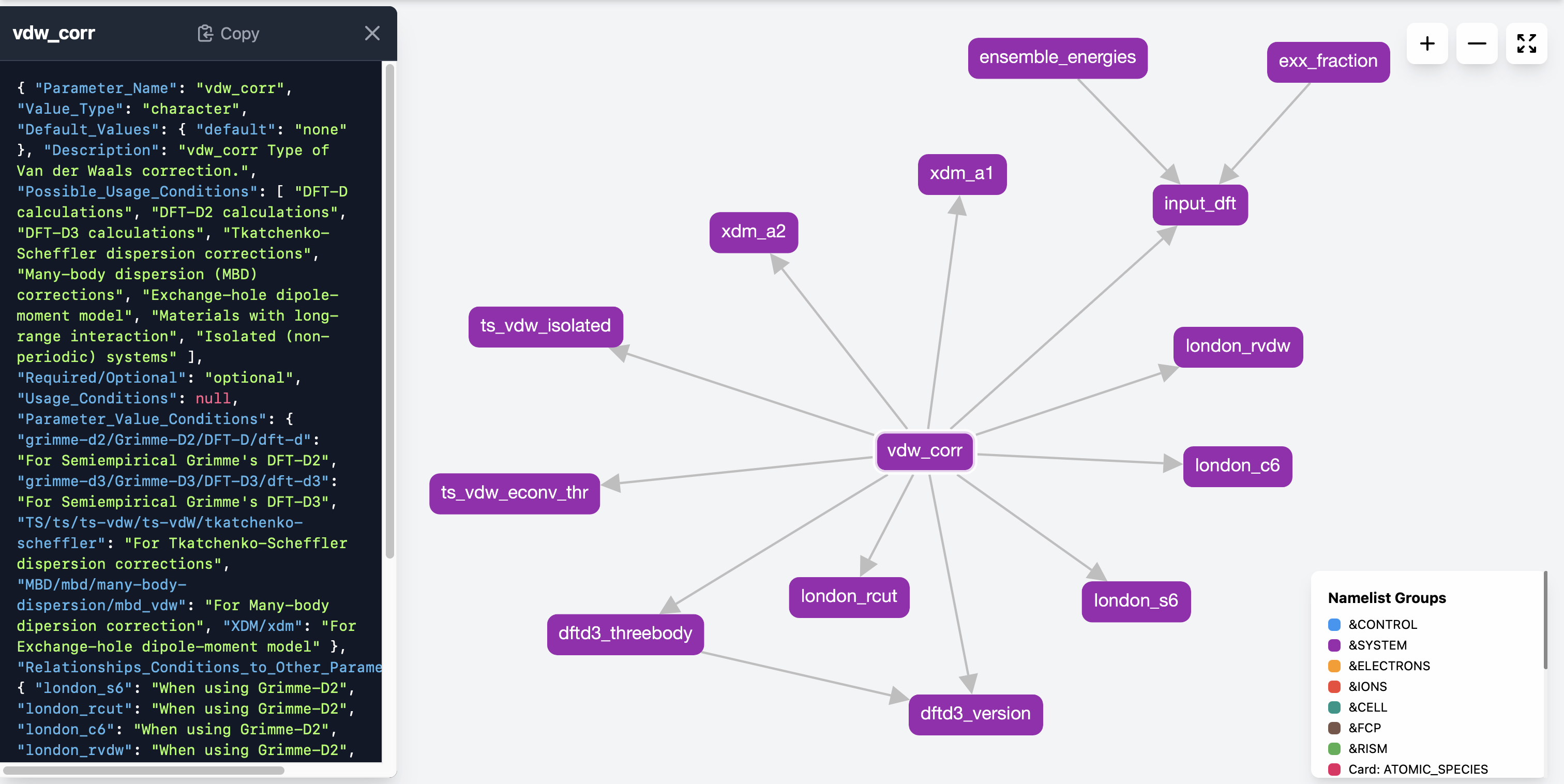}
        \caption{Screenshot of the web interface for the Quantum ESPRESSO knowledge graph, enabling interactive exploration of nodes and their relationships.}
        \label{fig:ep}
        \end{subfigure}
        \vspace{7pt}
    \begin{subfigure}[t]{0.49\textwidth}
        \centering
        \includegraphics[width=\textwidth]{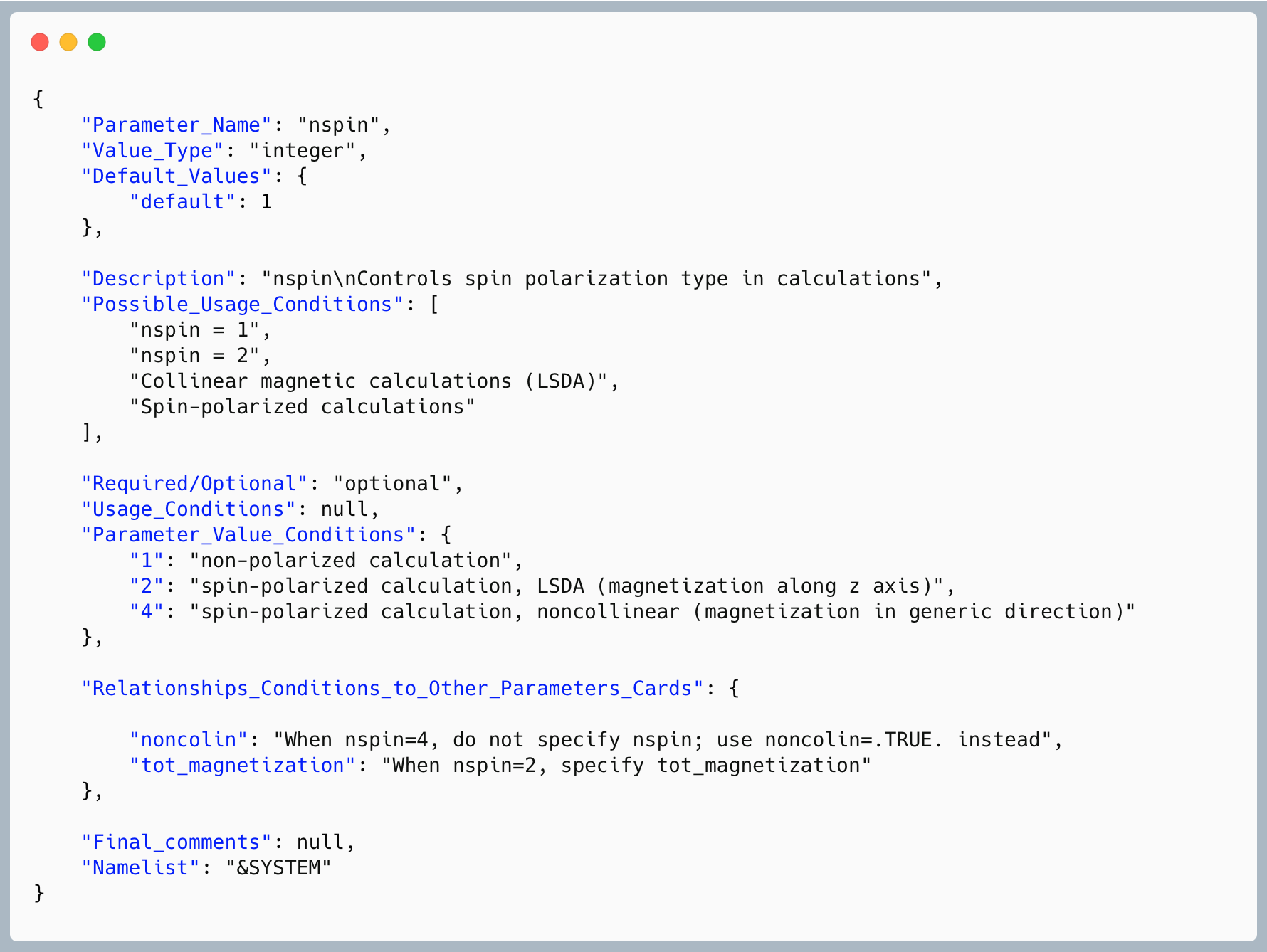}
        \caption{\parbox[t]{0.92\linewidth}{\raggedright Example \texttt{JSON} structure for a namelist parameter node \texttt{'nspin'} in the knowledge graph, showing its attributes and inter-parameter relationships.}}
        \label{fig:param}
    \end{subfigure}
    \hfill
    \begin{subfigure}[t]{0.49\textwidth}
        \centering
        \includegraphics[width=\textwidth]{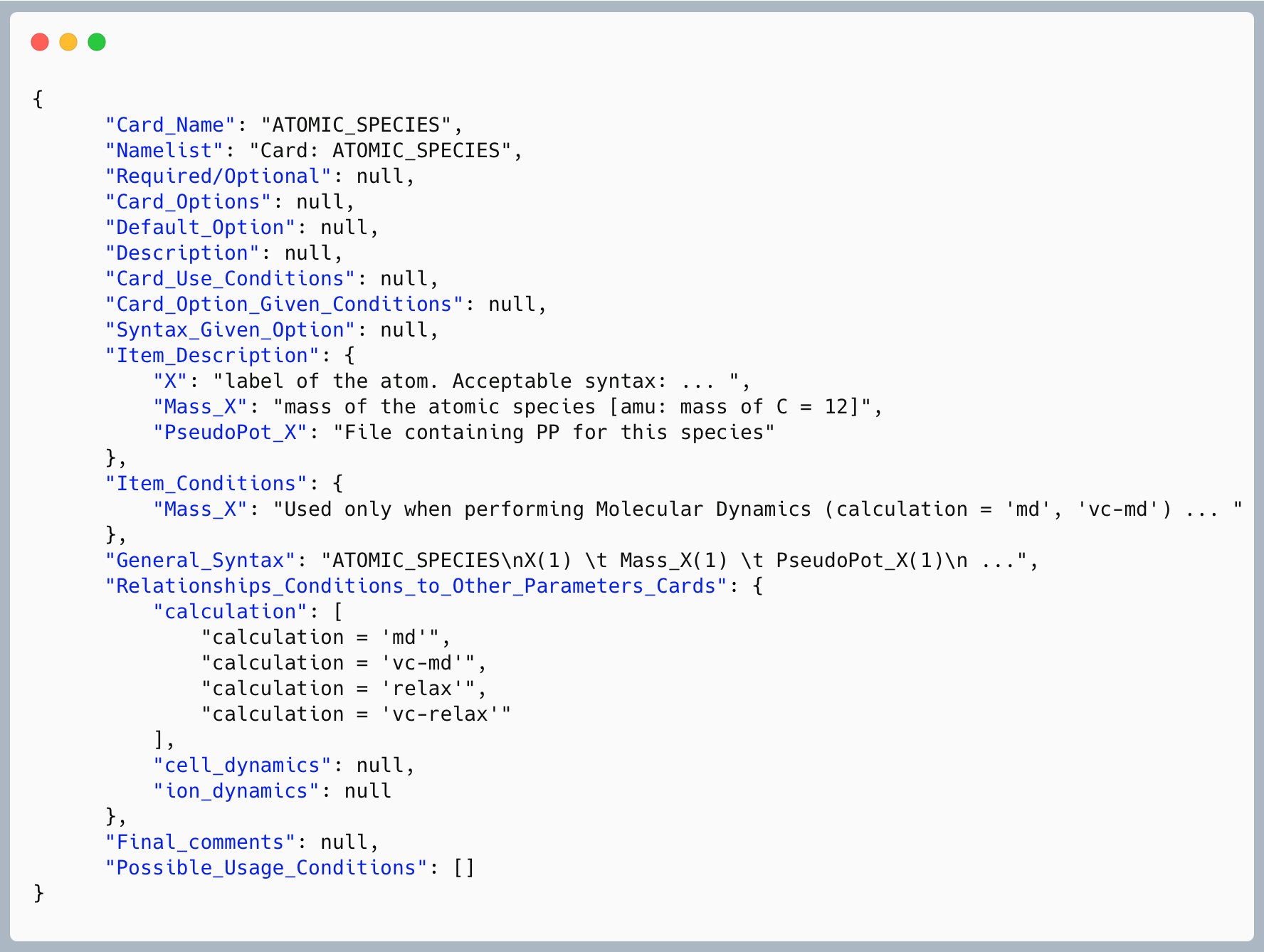}
        \caption{\parbox[t]{0.92\linewidth}{\raggedright Example \texttt{JSON} structure for a card node \texttt{'ATOMIC\_SPECIES'}, illustrating its complex syntax and conditional options.}}
        \label{fig:card}
    \end{subfigure}
    \caption{The three panels illustrate the structure and interface of the QE knowledge graph and how individual QE input parameters are formalized as machine-readable nodes and exposed through an interactive graph that makes their dependencies explicit.}
    \label{fig:kgprompt}
\end{figure}

The user interfaces with the original documentation, as provided, via our developed smart KG, as shown in \figref{fig:ep}, which displays the structure of the QE KG. The user can search the nodes and look for the connectivity between them. Although we performed some manual refinement, it should be noted that the parameter descriptions remain faithful to the original documentation. The KG is the recommendation system's core component, comprising \num{247} nodes and \num{330} connectivity edges, providing structured and connected information extracted from the QE program documentation. The data source for the QE knowledge graph is the online documentation for the QE \texttt{pw.x} code, which can be found at \url{https://www.quantum-espresso.org/Doc/INPUT_PW.html}. Initially, the documentation was converted into a plain \texttt{.txt} version, as QE uses parameters organized into \texttt{nameless} sections and \texttt{cards}, each with distinct syntax and purposes, which made this format easier to manage than the original \texttt{HTML} format. As a result, information was extracted separately for each type. The curated documentation was then transformed into a key-value pair format using \texttt{anthropic/claude-3.5-sonnet}\cite{claude-3.5-sonnet} language model, followed by manual verification and adjustments. For a complete description of the resulting key-value schema (including all namelist and card entries, their \texttt{connections} and \texttt{conditions} keys, and illustrative \texttt{JSON} examples such as the parameter \texttt{nspin} and card \texttt{ATOMIC\_SPECIES} given in \figref{fig:param}–\figref{fig:card}), please refer to our GitHub repository at \url{https://github.com/KIT-Workflows/agentic-workflow-framework/blob/main/knowledge_graph_schema.md}. Given the specialized expertise required, improving the KG, particularly the \emph{connections} and \emph{conditions}, remains an area for future work and potential community contributions. We refer to this object as \emph{smart} when we name it as \textit{smart} knowledge graph to distinguish it from a plain.

The nodes are retrieved from the KG using two complementary approaches: ($i$) direct keyword matching and ($ii$) context-aware retrieval based on graph relationships and inferred logical conditions. A threshold of top \SI{70}{\percent} cosine similarity\cite{Harald_2024} is used as the cutoff for the retrieval, increasing the number of nodes for evaluation, which means more computational resources are required. A hashing vectorizer is employed for node embedding, since it is robust, reliable, fast, and works well with large corpora. Due to the domain-specific nature of calculation requests, retrieval based on keywords and inferred conditions proved highly effective and efficient compared to denser embedding representations. Additionally, implicit conditions were inferred from the user's input request. For example, when a calculation mentions the \ce{Cu} surface, bulk, or elemental system, the \texttt{Metallic systems} condition is automatically invoked. We extracted \num{162} conditions under nine different categories to facilitate knowledge retrieval. These nine categories are: calculation type, functional and method, cell and material properties, pseudopotential, magnetism (and spin), isolated systems (by including boundary conditions), \textbf{k}-point settings, electric field conditions, and occupation types. User input requests are processed under these nine categories, and the explicit and implicit relevant conditions are extracted. Each condition serves as an access key to nodes in the KG. This inference of implicit conditions allows the KG to provide relevant information even when not explicitly requested, significantly enhancing the contextual understanding presented to the user or downstream tasks.  

\subsection*{Large Language Model Integration}
We employed LLMs to perform various tasks, including user prompt interfacing for material information, keyword extraction for KG retrieval, evaluation of QE parameters, and processing QE error messages. Users choose the LLM based on accuracy, cost, context window length, and runtime availability. For initial interfacing, condition extraction, and parameter evaluation, we used \texttt{mistralai/mixtral-8x22b-instruct}\cite{mixtral-8x22b-instruct}. For error keyword extraction, \texttt{databricks/dbrx-instruct}\cite{databricks_dbrx} was employed. Our core protocol generation involved a hierarchical model structure with two worker models (\texttt{databricks/dbrx-instruct}, \texttt{meta-llama/llama-3.1-405b-instruct})\cite{meta_llama3_1} and one referee model \texttt{anthropic/claude-3.5-sonnet}\cite{claude-3.5-sonnet}. Additionally, \texttt{google/gemini-2.0-flash-001}\cite{gemini-2.0-flash-001} was used for pre-processing user prompts to extract scoring metrics. This framework validates the effectiveness of the knowledge graph and the automated error-handling system, providing insights into LLM capabilities.

We implemented two main prompt engineering strategies. The first aimed to provide contextual scaffolding to the LLM, and the second strategy focused on extracting structured information. The first is used for subsequent LLM inputs and further processing or initiating other framework components; when generating textual responses, the LLM was guided to explain the current context and then expand upon it, making a reasoned conclusion based on incrementally acquired in-context knowledge. Such a technique was crucial for error handling and resolution, but less critical for tasks like keyword extraction. The second strategy, aimed at structured information extraction, employed explicit schema definitions with expected keys and data types, supported by few-shot examples, ensuring that the output was a valid \texttt{JSON} object. The LLM's raw output was extracted using regular expressions and parsed into a Python dictionary. This method helped to secure downstream integration for API calls and system updates.

\begin{figure}[h!]
    \centering
    \begin{subfigure}[t]{0.41\textwidth}
        \centering
        \includegraphics[width=\textwidth, height=7cm, keepaspectratio]{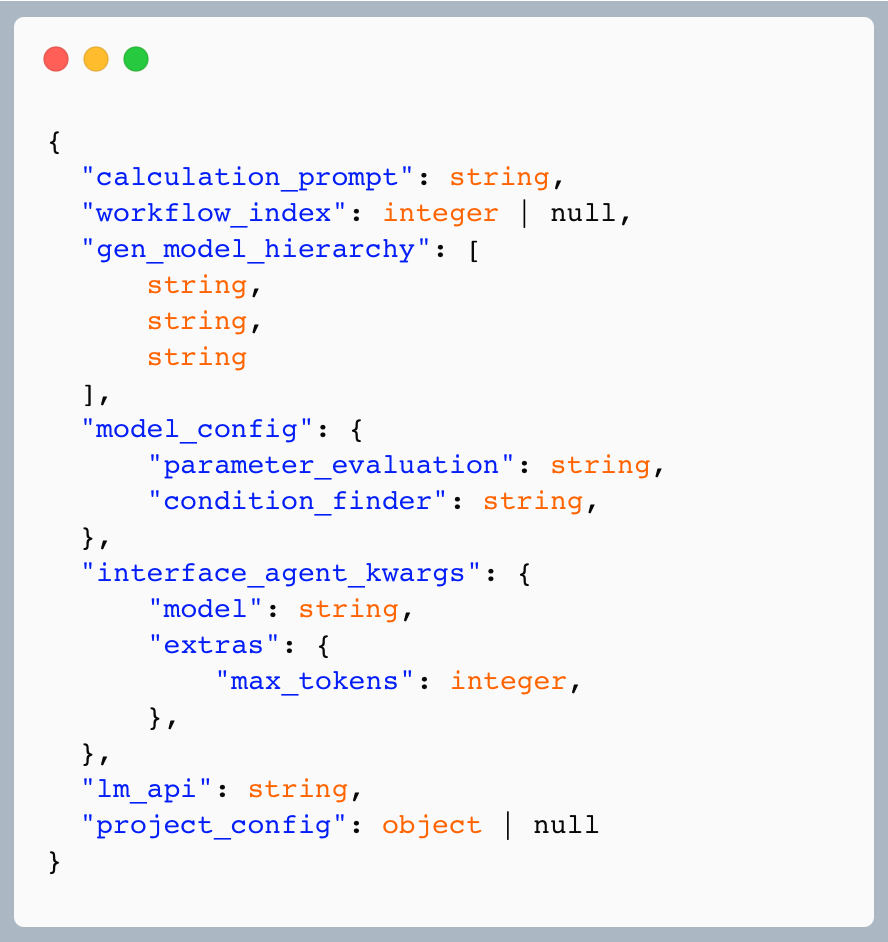}
        \caption{\parbox[t]{0.9\linewidth}{\raggedright Schematic of the minimal \texttt{JSON} payload accepted by the \texttt{POST /workflow/} REST endpoint.  The object defines the free-text \texttt{calculation\_prompt}, the ordered \texttt{gen\_model\_hierarchy}, per-task \texttt{model\_config}, optional interface-agent kwargs, the target LLM API, and an optional project configuration.}}
        \label{fig:jsload}
    \end{subfigure}
    \vspace{10pt}
    \begin{subfigure}[t]{0.58\textwidth}
        \centering
        \includegraphics[width=\textwidth, height=8cm, keepaspectratio]{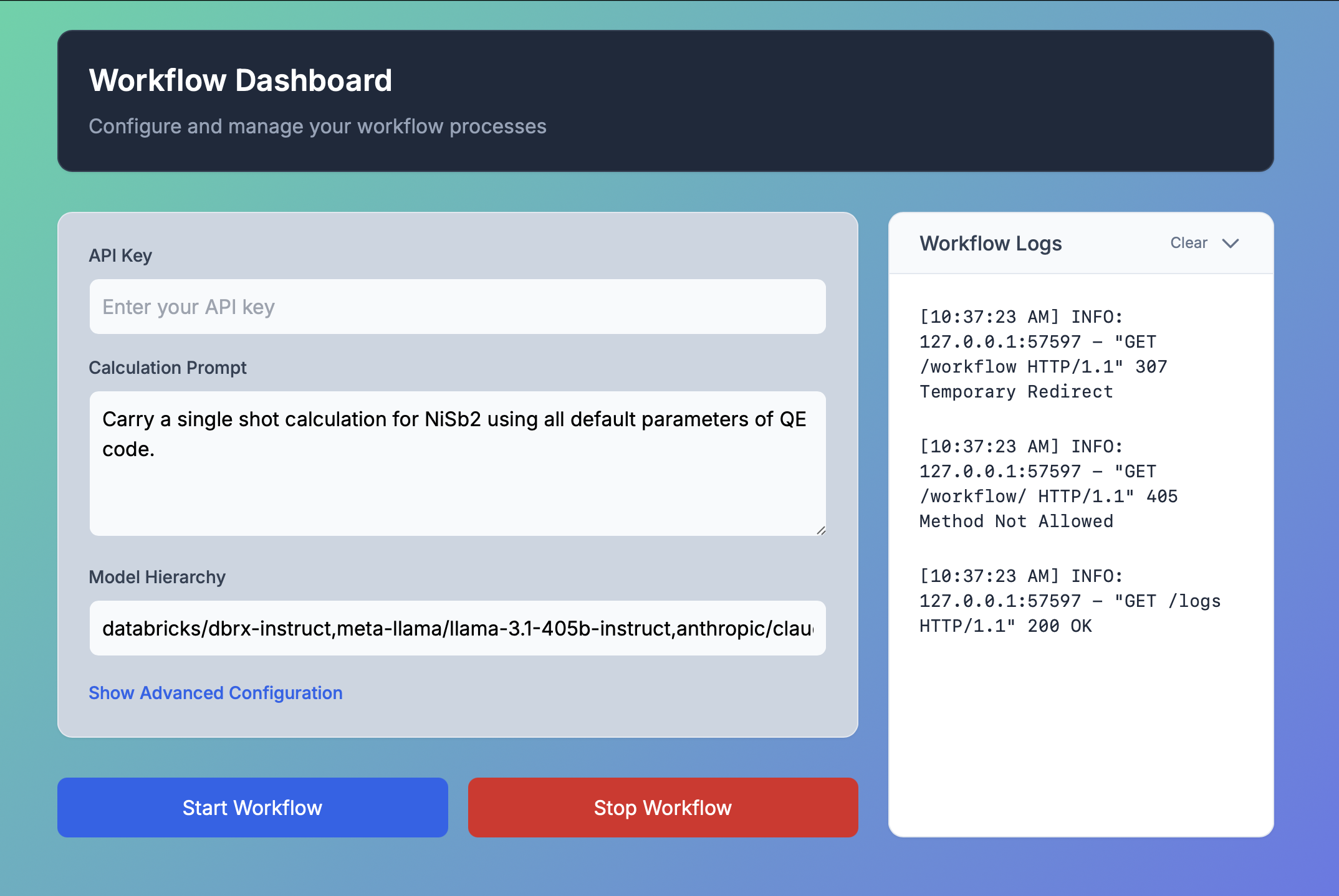}
        \caption{\parbox[t]{0.9\linewidth}{\raggedright The browser dashboard wraps the same parameters in a point-and-click interface: users paste an API key, enter their calculation prompt, select a model hierarchy, and launch or abort the run. A live log pane streams Server-Sent Events for real-time monitoring. The two views illustrate parity between programmatic and interactive control of the GENIUS workflow service.}}
        \label{fig:wfep}
    \end{subfigure}
    \caption{Comparison of workflow service JSON payload and the web interface.}
    \label{fig:fastapi}
\end{figure}

As displayed in panel \figref{fig:wfep}, user calculation prompts are parsed to extract keywords and calculation conditions. This structured output facilitates three parts of access to KG nodes. ($i$) The required nodes corresponding to essential QE parameters for any calculation are included. ($ii$) A keyword search is performed using the extracted keywords against the raw text of the knowledge graph. The top \SI{70}{\percent} of nodes ranked by relevance (cosine similarity) are selected. This text search utilizes a hashing vectorizer (\texttt{scikit-learn} implementation) with $2^{17}$ features to vectorize the raw text in the knowledge graph and the query keywords. ($iii$) QE-specific conditions are extracted from the user's prompt. We identified \num{164} unique conditions across nine categories within the QE documentation. The user's prompt is parsed to identify applicable conditions within these categories. Each matched condition activates relevant KG nodes. The nodes connected to the collected nodes are added to the final set. The nodes from these three parts are concatenated and sent for evaluation. Each selected node (QE parameter) is assessed based on the user's prompt and calculation conditions to determine an appropriate value. If a parameter is evaluated as non-relevant, it gets \texttt{None} as a value and is excluded.

\subsection*{Prompt Dataset and Complexity Mapping} 
As depicted in \figref{fig:wfep}, the framework provides a web interface through which users submit free-form calculation prompts. Moreover, since material-structure geometries can be inconsistent when directly extracted from different models, we employ an agent responsible for retrieving and standardizing all structural data from the specified database to ensure the reproducibility of the DFT calculations. Upon submission, the system parses the text, extracts the material formula or structure, and automatically routes the request to the appropriate Materials Cloud database (MC2D or MC3D) based on whether the target compound is two- or three-dimensional \cite{huber2022materials,campi2023expansion}. To benchmark the interface under realistic conditions, we assembled more than 400 prompts from independent researchers without prior knowledge of the framework’s internal design. After verifying that each requested material was available in MC2D or MC3D, we selected \num{295} prompts for testing the framework. We quantified the prompt complexity with a score-based rubric inspired by Brown \textit{et al.}\cite{brown2020language}. Using the \texttt{google/gemini-2.0-flash-001} model, we extracted ten linguistic and domain-specific features from every prompt; each feature present assigned $+1$ to the total score. Prompts scoring $0\text{–}4$, $5\text{–}8$, and $9$ or more were labeled basic, standard, and complex, respectively. The complete scoring scripts and prompt dataset are available in our public \href{https://github.com/KIT-Workflows/agentic-workflow-framework}{repository}\cite{agentic-workflow-framework} for more details.

\subsection*{Automated Error Handling (\texttt{AEH})}
The framework incorporates an automated error handling (\texttt{AEH}) system designed to mitigate the critical challenge of time-consuming manual debugging when simulations fail. Following the generation of a simulation protocol, the QE program is executed. If the execution fails, QE generates a \texttt{CRASH} file containing a descriptive error message. An LLM processes this message to extract relevant keywords, which are then used to query the smart KG for relevant information. The retrieved nodes provide contextual information to the LLM, which then attempts to formulate a solution to the encountered error. Each LLM within the hierarchical architecture is allocated a predefined number of retries. Logs from previous unsuccessful attempts within the same model's retry cycle are omitted from the LLM input due to their length and potential lack of immediate usefulness. If an LLM uses all its retries without resolving the error, it switches to the next model in the hierarchy. The subsequent model restarts the error resolution process from the beginning, using the initial output from the recommendation module as its starting point. The overall effectiveness of the \texttt{AEH} system is thus significantly dependent on the clarity and informational content of the \texttt{CRASH} file combined with the KG, operating within a self-consistent feedback loop. The QE \texttt{PWSCF v.7.2} package\cite{QEpwguide} serves as the execution-level validator for the generated simulation protocols. A protocol is considered valid if the QE simulation completes successfully, as indicated by the absence of a \texttt{CRASH} file and a zero exit code. Conversely, an invalid protocol results in a non-zero exit code and the generation of a \texttt{CRASH} file that contains a detailed error description.

Logs were collected from executing the $295$ curated calculation prompts to track the framework's evolution. Each log file records the terminal status of the workflow (success or failure) and, in the case of successful runs, includes the number of attempts and any model switches that occurred during the automated error-handling phase. The primary metric for assessing the efficiency of protocol generation is the number of attempts and model switches required to achieve a successful outcome. A zero value indicates zero-shot generation, indicating that the initial protocol generated by Model 1, using the recommendation system's output, was successful on the first attempt, without requiring any intervention from the \texttt{AEH} system. The primary requirements are the capability to interact with LLM provider APIs and to execute the QE program for protocol validation. Some tools, such as the Atomic Simulation Environment (ASE), \texttt{scikit-learn}, and \texttt{networkx}, were also used. Additional dependencies are standard Python libraries. The workflow API is a \texttt{RESTful} design written in \texttt{FastAPI}, managing the workflow through standardized \texttt{HTTP} endpoints. Once the workflow server is initialized, its primary endpoint \texttt{POST "/workflow/"} accepts \texttt{JSON} payloads that specify calculation parameters, model configurations, and optional project settings, as shown in \figref{fig:jsload}. The service provides additional workflow management through other endpoints for: Status monitoring (\texttt{GET "/workflow-status/\{workflow\_id\}"}), Result retrieval (\texttt{GET "/results/\{workflow\_id\}"}), and Visualization access (\texttt{GET "/timeline/\{workflow\_id\}"}). Real-time monitoring is enabled via Server-Sent Events (SSE) at the \texttt{/logs} endpoint, as illustrated in \figref{fig:wfep}. These approaches facilitate programmatic integration with other AI-driven applications while supporting interactive human user experiences.

\section*{Results}

\begin{figure}[h!]
    \centering
    \resizebox{0.99\textwidth}{!}{%
        \includegraphics{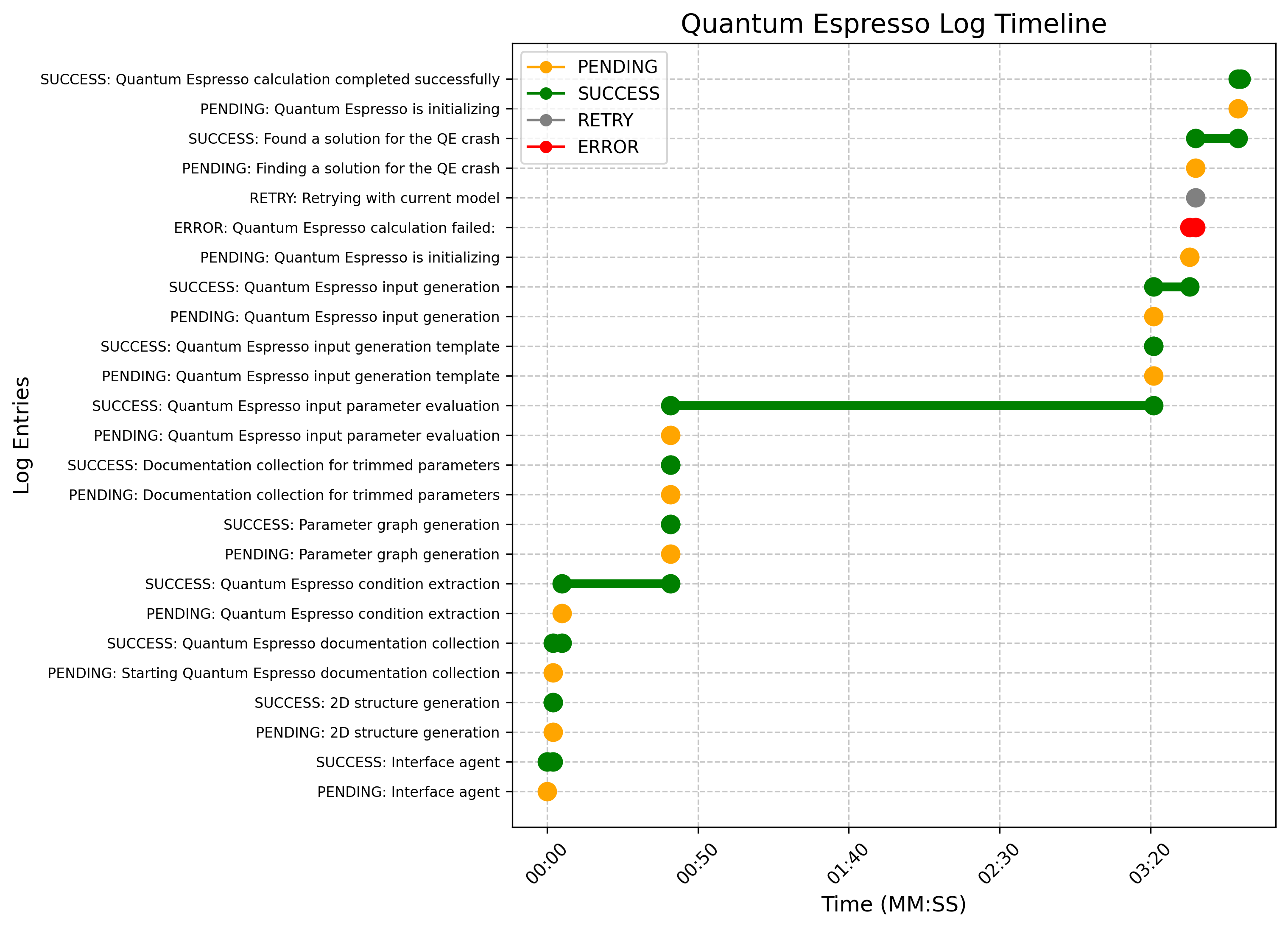}
    }
    \caption{Live timeline of a self-healing (\texttt{AEH}) GENIUS job. Each dot marks a log event (y-axis, newest at top) plotted against wall-clock time (x-axis), colors denote status: \texttt{PENDING} (orange), \texttt{SUCCESS} (green), \texttt{RETRY} (gray), \texttt{ERROR} (red). The workflow first parses the user prompt, harvests documentation, builds a parameter graph, and generates a QE input template. After launch, QE crashes once (red); the finite-state loop applies a single retry (gray) within the \texttt{AEH}, resolves the issue, and the simulation reaches steady execution and completion (green) in $\approx$3 min. The timeline exposes full provenance and illustrates how GENIUS autonomously recovers from runtime failures while streaming real-time status updates.}
    \label{fig:qe-log}
\end{figure}

We tested GENIUS with multiple LLMs to evaluate our approach, each exhibiting incremental capabilities, to obtain more comprehensive diagnostic insights into the workflow's performance. By gathering a large dataset of human-generated prompts for DFT calculations through QE and collecting the corresponding workflow logs (see \figref{fig:qe-log}), we analyzed how many attempts were required to complete each prompt successfully. These prompts were authored by chemists and physicists who routinely perform DFT simulations with electronic-structure packages other than Quantum ESPRESSO, ensuring that the benchmark reflects realistic expert usage while remaining unbiased toward QE-specific syntax.

To understand the diversity of the prompts dataset, we converted the \num{295} prompts into \num{3072}-dimensional embedding vectors with OpenAI’s \texttt{text-embedding-3-large} model. A $10 \times 10$ self-organizing map (SOM) \cite{kohonen1990self} was then trained to visualize their semantic landscape. The SOM is an unsupervised neural network for dimensionality reduction and clustering by projecting high-dimensional input data onto a lower-dimensional grid (here \num{2}-dimensional) while preserving topological relationships between the input data. Each of the \num{100} SOM neurons was initialized with a random weight vector of equal dimensionality, and training proceeded for \num{50000} iterations in mini-batches of \num{50} samples. During each iteration, the neurons compete to best represent the input pattern. The algorithm identifies the Best Matching Unit (BMU) for each input vector, which is defined as the neuron whose weight vector has the smallest Euclidean distance to the input vector. Afterward, the BMU and its neighboring neurons are updated, with the adjustment magnitude decreasing as a function of distance from the BMU, according to a Gaussian neighborhood function. This neighborhood influence, combined with a linearly decreasing learning rate, allows the SOM to form a topologically ordered map of the input space. The convergence and quality of the resulting map were quantitatively validated by calculating standard SOM metrics: the Quantization Error, representing the average distance between input data vectors and their best matching unit's weight vector, and the Topological Error (TE), measuring the proportion of data points for which the first and second BMUs are not adjacent on the map grid.

\begin{figure}[t]
    \centering
    \begin{subfigure}[t]{0.49\textwidth}
        \centering
        \includegraphics[width=\textwidth, keepaspectratio]{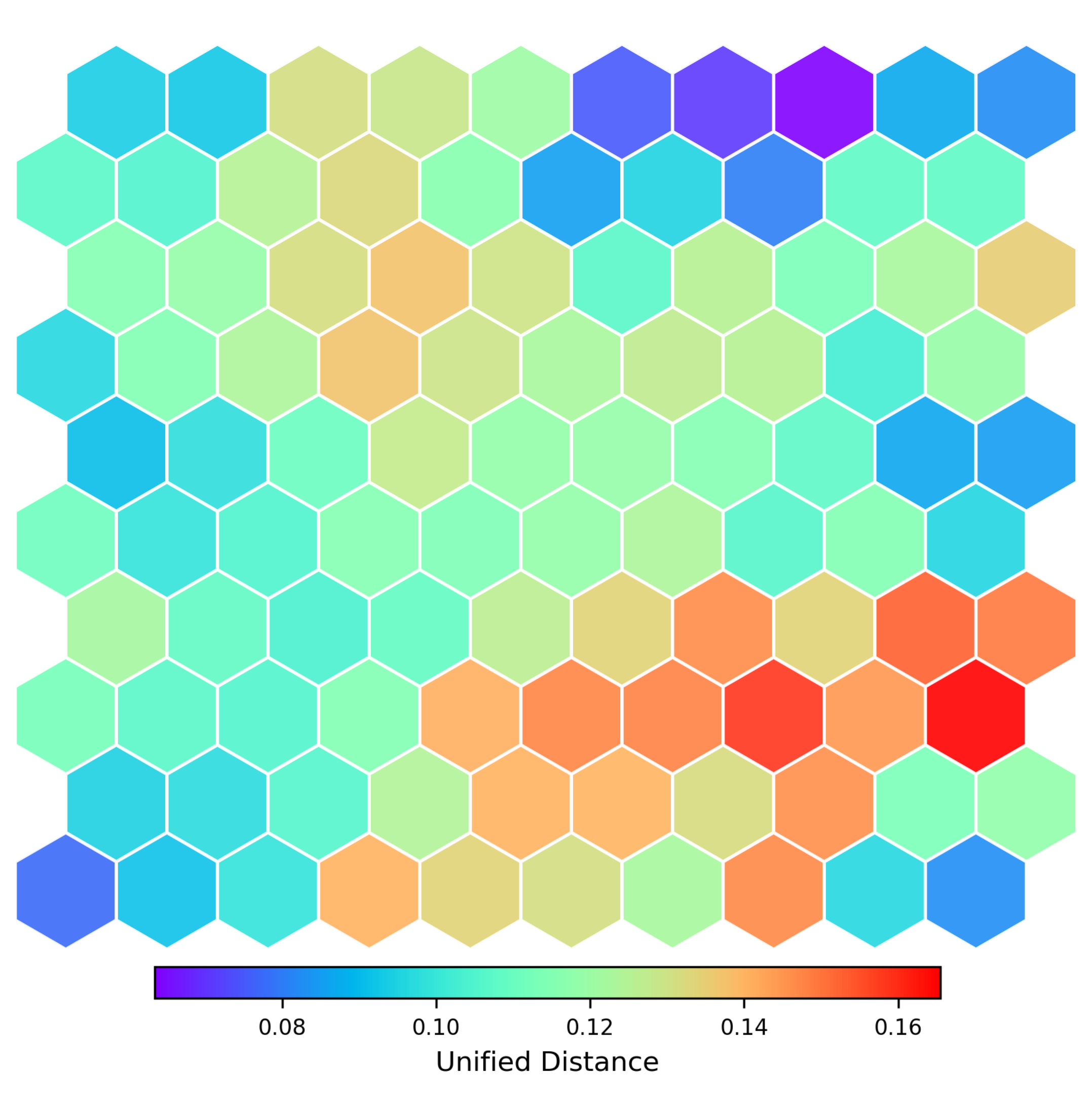}
        \caption{\parbox[t]{0.92\linewidth}{\raggedright U-matrix visualization of the SOM trained on user prompt embeddings, where regions with more distance indicate cluster boundaries and regions with lower distance denote dense clusters of similar prompts.}}
        \label{fig:somu}
    \end{subfigure}
    \vspace{10pt}
    \begin{subfigure}[t]{0.49\textwidth}
        \centering
        \includegraphics[width=\textwidth, keepaspectratio]{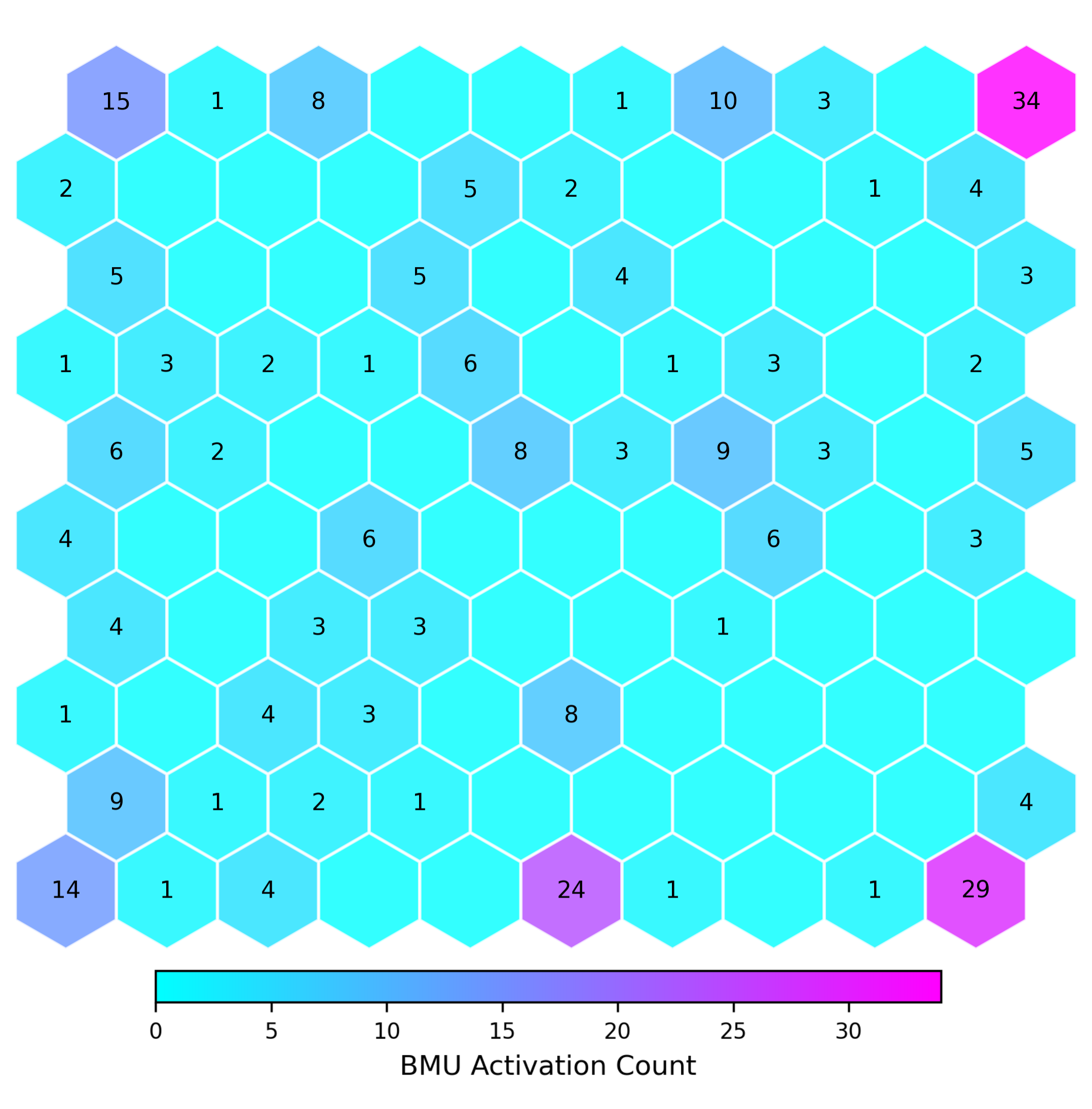}
        \caption{\parbox[t]{0.92\linewidth}{\raggedright SOM hit map (BMU activation count) showing the distribution of prompts across the neuron grid. Five distinct high-activation and low-activation regions are scattered in between, suggesting a balance of semantic clusters and diverse request types in the input data.}}
        \label{fig:somd}
    \end{subfigure}

    \caption{Self-Organizing Map (SOM) analysis of user input prompt embeddings.}
    \label{fig:soms}
\end{figure}

The SOM and BMU analyses give us information regarding the prompts' degree of complexity and semantic similarity. The score-based metric evaluation shows that the prompts comprise \SI{44.3}{\percent} basic, \SI{48.5}{\percent}  standard, and \SI{7.2}{\percent} complex prompts. This evaluation is performed by an LLM, which means that the language model assigns a numerical value to text data\cite{brown2020language}. The SOM analysis creates a self-organizing map grid of hexagonally packed neurons, \figref{fig:soms}, trained on the embeddings of the prompts, and the prompts are clustered into different groups. The SOM shows the clusters while preserving topological identity. The quality of the SOM representation reinforces the reliability of the observed prompt distribution (\figref{fig:soms}). The low TE (\num{0.0373}) confirms excellent preservation of the original data's neighborhood structure. The Quantization Error ($0.4970$) indicates good representational fidelity; considering that the input vectors were unit-normalized (maximum possible pairwise distance of 2.0), this average distance between data points and their map representatives is low, especially given the significant dimensionality reduction. The U-matrix (Unified distance matrix) in \figref{fig:somu} shows the neuron distances; the hexagonal structure captures six equidistant neighbors, compared to a square grid with only four. To analyze the learned representations, two complementary visualizations were generated. The U-matrix visualizes the average distance between neighboring neurons, where higher values indicate cluster boundaries and lower values suggest dense regions of similar inputs. The BMU activation count plot (hit map) in \figref{fig:somd} shows how frequently each neuron was selected as a BMU, revealing the distribution of input assignments across the SOM grid. Neurons with zero activations indicate they serve as boundary regions or represent semantic areas not covered by the current dataset. These \emph{empty} neurons are crucial for topology preservation, helping maintain proper distance relationships between clusters.

A similar SOM representation can be generated for each component of the embedding vectors, revealing which dimensions contribute most significantly to semantic distinctions and helping identify correlated components that form meaningful semantic features. These visualizations are available in the project's GitHub \href{https://github.com/KIT-Workflows/agentic-workflow-framework}{repository}. The figure shows five neurons with the highest activation counts, including their respective BMUs. These neurons are well separated on the SOM grid, indicating that the calculation prompts can be grouped mainly into five semantic clusters, with additional residual prompts that share similarities within the core concepts. Overall, the SOM grid shows how the prompt collection is dispersed, which is a good balance between semantic cohesion within the identified clusters and conceptual diversity across them. Upon manual inspection of prompts, it was found that the prompts are mainly divided into two major categories, namely structural relaxation and a variety of single-shot DFT calculations, using different methodologies available in the QE code, which is further confirmed by a simpler k-means analysis (See GitHub \href{https://github.com/KIT-Workflows/agentic-workflow-framework}{repository}). Discussing these findings illuminates the framework's sensitivity to input quality and type, potentially guiding future improvements in user interaction or prompt pre-processing. This analysis uncovers latent structures within the user request space directly relevant to developing automated protocol generation mechanisms.

\begin{figure}[h!]
    \centering
    \resizebox{0.8\textwidth}{!}{%
        \includegraphics{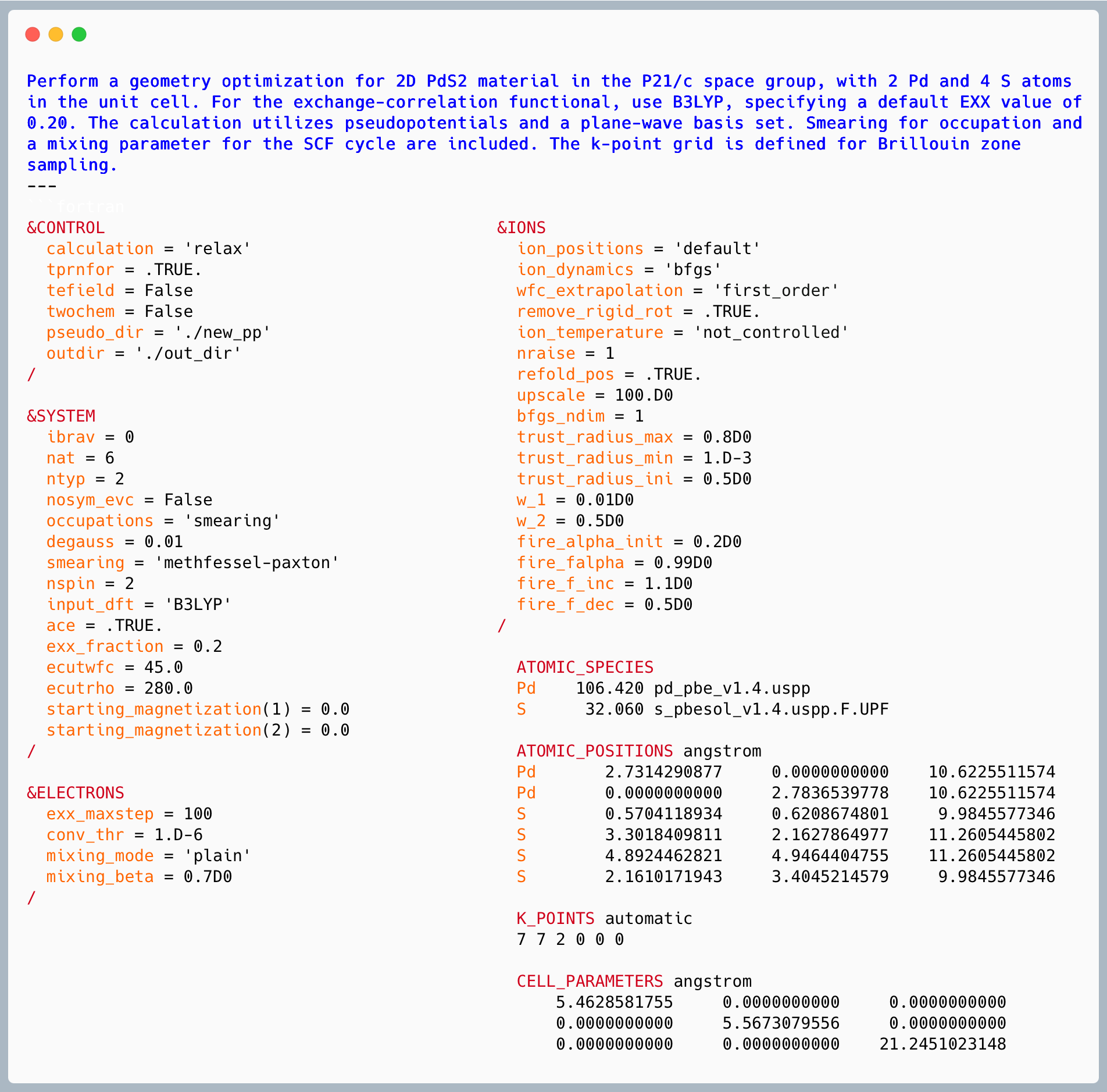}
    }
    \caption{Real simulation protocol example of Quantum Espresso generated by GENIUS. The user's prompt request is displayed in the upper part, instructing QE code to perform a geometry optimization for 2D \ce{PdS2} in the P21/c space group, using Quantum Espresso with the \emph{B3LYP} exchange-correlation functional. The generated protocol is provided, as presented in two columns for compactness. The framework parses these instructions and automatically generates the valid QE input. The protocol specifies \SI{20}{\percent} exact-exchange, a plane-wave basis set, smearing for occupation, a mixing parameter for the SCF cycle, and a $7\times7\times2$ \textbf{k}-points mesh. Additionally, the file includes detailed control parameters for geometry relaxation (via BFGS), pseudopotentials for \ce{Pd} and \ce{S}, and the required settings, such as ecutwfc, ecutrho, occupations, spin polarization, as presented in the output. }
    \label{fig:qe-in}
\end{figure}

\begin{figure}[t]
    \centering
    \resizebox{1.1\textwidth}{!}{%
        \includegraphics{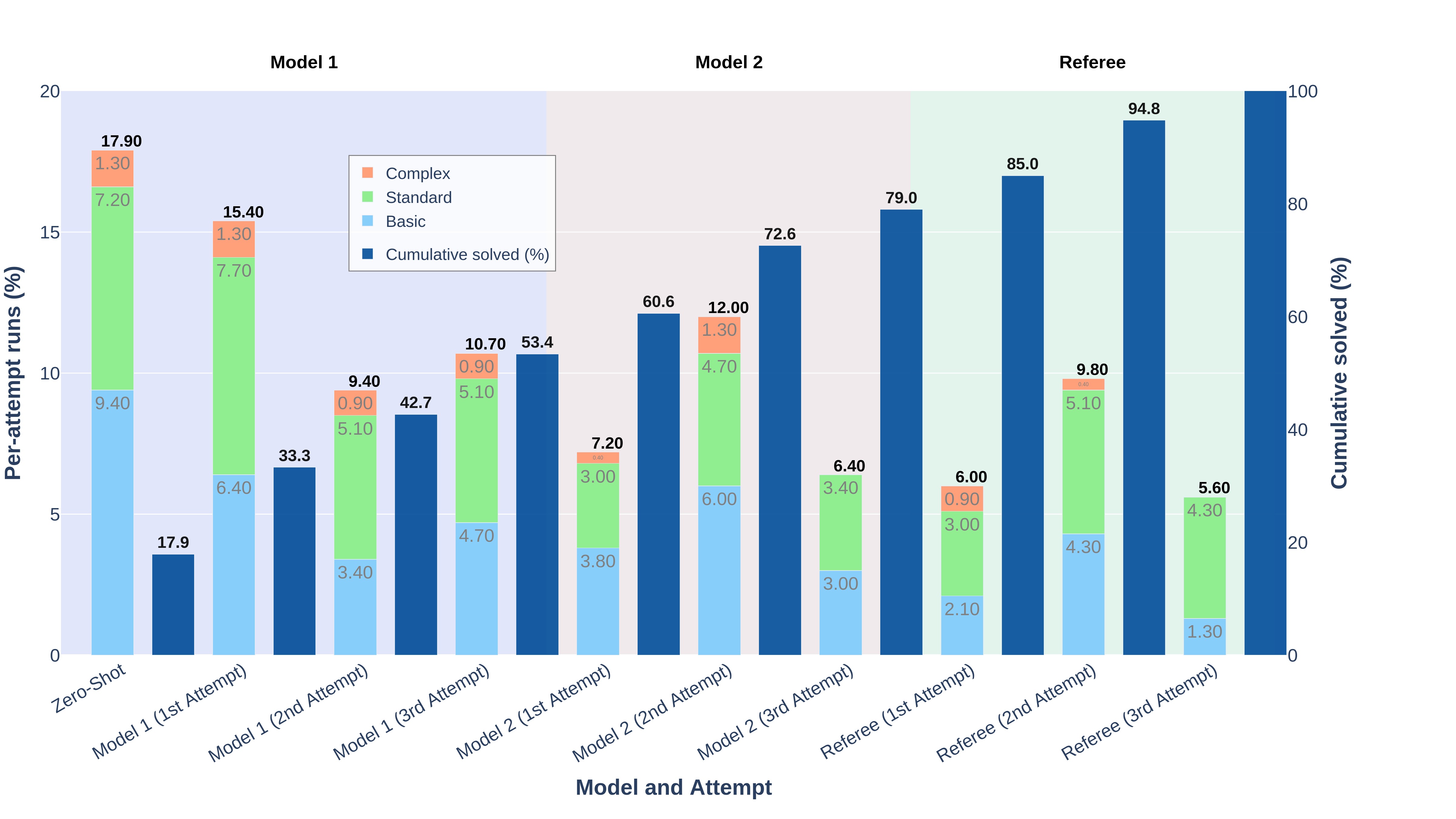}
    }
    \caption{GENIUS performance benchmark on \num{295} tested prompts.
The stacked bar chart reports the \emph{percentage of successful runs} (y-axis) for the \emph{Zero-Shot}  pass (GENIUS without \texttt{AEH}) and GENIUS using \texttt{AEH} combined with \emph{Model 1}, \emph{Model 2}, and \emph{Referee} models. Shaded vertical panels group the bars that belong to the systems assessed by the same model. Within every bar, colored segments disaggregate the total success rate by prompt complexity: Basic, Standard, and Complex. The cumulative solved percentage (right y-axis) is overlaid in dark blue, showing the total proportion of prompts solved after each successive attempt.
}
    \label{fig-bars}
\end{figure}

An overview of the outcomes for the \num{295} test prompts is presented in \figref{fig-bars}, which depicts the distribution between successful and failed runs, the path to success, \emph{zero-shot} or via specific models in the \texttt{AEH} system, as well as a breakdown according to the initial prompt complexity. Additionally, when the prompts are evaluated using only base LLMs without the GENIUS framework, they return negligible contributions to generate valid QE input files containing the correct cards and mutually consistent parameters for a given geometric structure, irrespective of their nominal reasoning enhancements. This limitation is probably because the models do not embed explicit crystallographic information and cannot infer the subtle interdependencies between geometry, namelist keywords, and the card syntax required by QE. We reiterate that Model 1, the first component in the protocol generation hierarchy, produces the initial simulation protocol version. Therefore, a \emph{zero-shot} success corresponds to cases in which the first output generated by Model 1 is valid and does not require any further correction. Within the GENIUS framework, Model 1 proceeds with the first cycle of automated error-handling retries if this initial attempt fails.

In \figref{fig:qe-in}, the user's prompt (classified as standard) is shown at the top, specifying a geometry optimization for a \ce{2D} \ce{PdS2} structure using the \texttt{B3LYP} functional, whereas the bottom portion highlights the valid QE input file generated by the GENIUS framework. In \figref{fig:qe-log}, we illustrate the complete timeline log for the same prompt, emphasizing the sequence of events, as indicated by color-coded statuses (\texttt{PENDING}, \texttt{SUCCESS}, \texttt{RETRY}, and \texttt{ERROR}), as the system progresses from the interface agent phase to the final solution generation. As evidenced in \figref{fig:qe-log}, the extended time required to evaluate input parameters is a secondary constraint imposed by LLM API providers, which a self-host service could mitigate; without such limitations, parameters could be processed in parallel, reducing overall latency. The stepwise entries in the log figure demonstrate the framework's resilience, including how QE crashes (red dots) occur due to some hallucination or confabulation. Our framework automatically detects and resolves these failures, which iteratively refines and validates the input parameters until the final QE calculation is completed.

In \figref{fig-bars}, we present the distribution of successful runs that reached the \texttt{FINISHED} state in the workflow after a given number of attempts. The success at zero-shot cases indicates the scenarios where a request was \texttt{FINISHED} using only the recommendation system of the workflow, without invoking the automated error handling system. This scenario accounts for \SI{17.9}{\percent}, comprising \SI{9.4}{\percent} for basic prompts, \SI{7.2}{\percent} for standard prompts, and \SI{1.3}{\percent} for complex prompts. A similar distribution pattern can be observed for subsequent attempts. If the initial execution fails, the GENIUS \texttt{AEH} system is triggered. Each retry uses the same model that generated the initial protocol within its designated attempt cycle. After three attempts per model, the process switches to the next model in the hierarchy if no solution is found. Based on the user's calculation prompt, the workflow resets from the output of the recommendation system, which is a template for generating a simulation protocol. The previous changelog attempts are not provided to the new model in each model exchange. The model receives only the error message, the relevant documentation, the latest version of the simulation protocol, and the original user calculation prompt.

Our results demonstrate that successful cases at each attempt comprise a mixture of basic, standard, and complex calculation prompts. This observation shows that prompt complexity (basic, standard, or complex) is not inherently problematic for the framework's performance. Complex prompts can contain more distinctive instructions, enhancing the framework's ability to generate valid protocols. The general trend reveals that after the initial attempts with Model 1, the number of successful attempts stabilizes at a baseline level. This initial high success rate is followed by a plateau, which resembles an exponential decay behavior in the number of cases requiring successive attempts. For the model selection hierarchy, we assume a performance ordering of $\text{Model } 1 < \text{Model } 2 < \text{Referee}$. This can be done with any set of language models, but the predefined order characterizes the framework's behavior, where the Referee model was chosen as the SOTA model. The Referee model is used to establish the existence of the performance baseline. This outcome indicates that the framework itself, rather than just the power of the strongest model, is responsible for successfully handling most cases, as the Referee model is not utilized disproportionately, which would otherwise suggest a failure in the preceding stages. This demonstrates that the GENIUS framework can be used with any model (it is model-agnostic) and that its overall performance is attributable to its architecture intelligence, not just the underlying language model's capabilities. The opposite scenario would manifest as a lack of a baseline and, instead, an increase in successful attempts. Specifically indicating that success relied primarily on the more performative model rather than the framework's architectural design.

From our total dataset of \num{295} calculation requests analyzed, \num{235} successfully produced a valid simulation protocol, with \num{42} of these succeeding in the \emph{zero-shot} scenario, which is defined here as the framework converging to a correct protocol on its very first attempt, without invoking any automated error-handling loops. This yields an overall system success ratio of $P(S) = \frac{235}{295} \approx 0.7966$, a 
\emph{zero-shot}, ($ZS$), success ratio of $P(ZS) = \frac{42}{295} \approx 0.1424$, and a ratio of success through automated error handling (given zero-shot fails) of $P(\texttt{AEH} \mid \text{not } ZS) = \frac{193}{253} \approx 0.7628$. To characterize how the success rate evolves as a function of successive attempts, we fitted an exponential decay function to the observed success rates ($S$) across multiple attempts, where $x$ represents the attempt number. The function takes the form, Eq. 
\ref{eq:dec}:
\begin{equation}
S(x) = A \, e^{-b x} + C,\quad\mathrm{RMSE} = 1.9\%~,
\label{eq:dec}
\end{equation}
obtaining $A = \SI{11.1}{\percent} \pm 1.0$, $b = 0.46\ (1/\mathrm{attempt}) \pm 0.1$, and $C = \SI{7.0}{\percent} \pm 0.70$. In this parametrization, the initial amplitude, $A\ (\si{\percent})$, represents the maximum influence of the zero-shot attempt; the decay rate, $b\ (1/\mathrm{attempt})$, determines how quickly this initial advantage decreases over successive attempts, and the baseline, $C\ (\%)$ is the asymptotic success probability reached after many retries. 

\begin{figure}[t]
    \centering
    \resizebox{1.0\textwidth}{!}{%
        \includegraphics{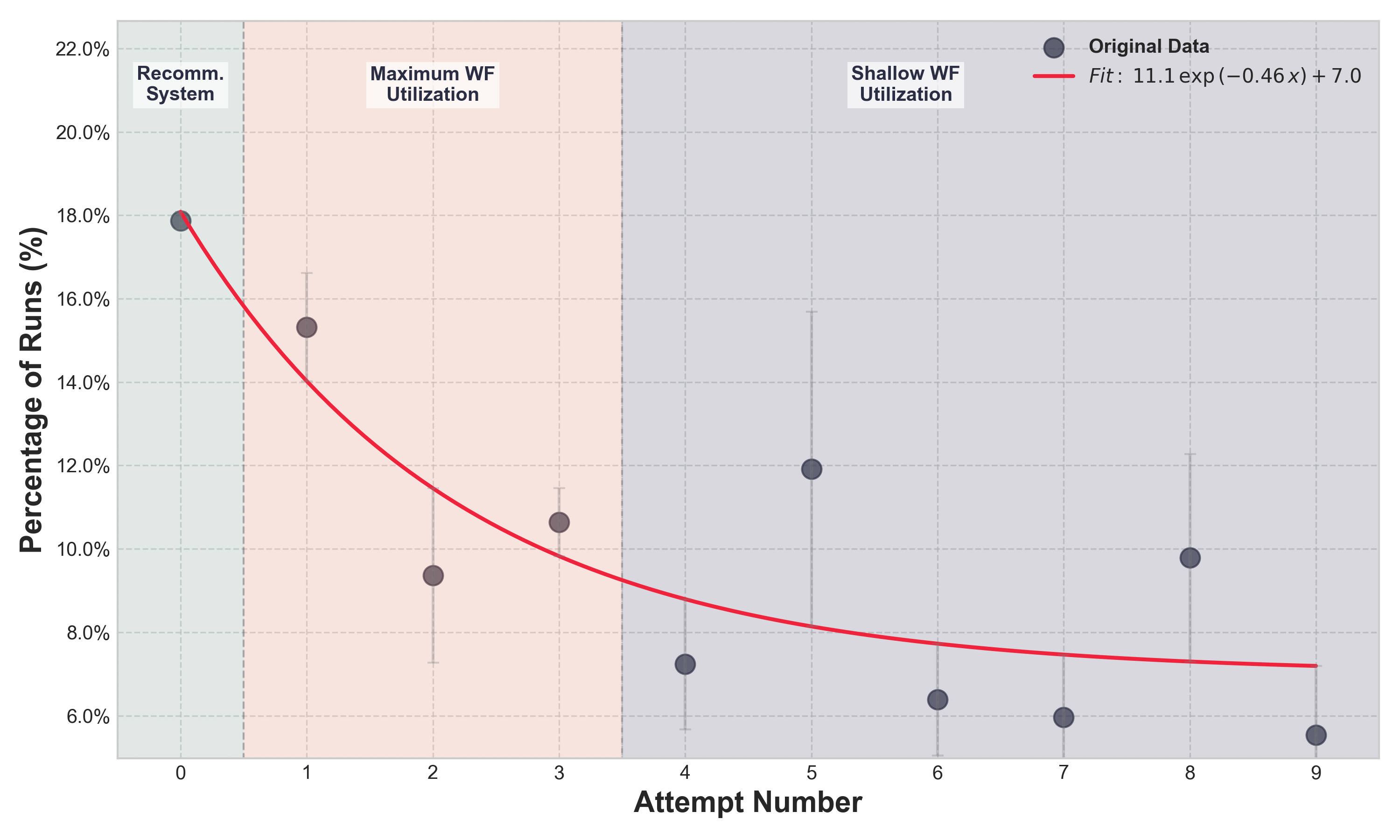}
    }
    \caption{Exponential decay fit (red curve) applied to the observed fraction of successful runs per attempt number (black points). A single-parameter exponential, $S(x)=11.1\,\mathrm{e}^{-0.46x}+7.0$ \% (red line), captures the trend. Shaded bands specify the three operating regimes: the opening \textbf{Recommendation System} zone (\texttt{zero-shot} wins), the steep \textbf{Maximum Workflow Utilization} zone where early retries yield rapidly diminishing but still substantive gains, and the long-tail \textbf{Shallow Workflow Utilization} zone in which performance plateaus at the \SI{7}{\percent} baseline.  The fit confirms that most recoverable errors are corrected within the first three attempts, after which additional computation yields marginal returns.}
    \label{fig-exp}
\end{figure}

\figref{fig-exp} delineates three distinct operational regimes within GENIUS: \textbf{recommendation-system}, \textbf{maximum workflow utilization}, and \textbf{shallow workflow utilization}. The opening \textbf{recommendation-system} regime coincides with the \emph{zero-shot} pass, highlighting the framework’s ability to successfully generate protocols independently of model switching or fallback mechanisms. Immediately thereafter, the curve plunges through the \textbf{maximum workflow utilization} regime: each early retry unlocks deeper cross-model synergies, yielding rapidly diminishing, but still substantive, gains. Once the process reaches roughly six attempts, the trajectory flattens into the \textbf{shallow workflow utilization} regime, where the performance asymptotically converges toward the baseline value of $C\!\approx\!7\%$. Within this regime, further retries contribute marginal benefit; success is governed primarily by the workflow’s inherent competence rather than by additional computation. The slight oscillations superimposed on the fitted curve stem from the design choice to reset the context after every third attempt and switch the model. Inter-block influence is shortened because each reset isolates the subsequent block of attempts. Allowing more consecutive attempts per model would amplify these oscillations, which could be quantitatively captured by extending the fitting function to include an explicit periodic component, where finer-grained modeling is required.

The recommendation system ($Rec$) includes the smart knowledge graph, extracts boundary conditions, and evaluates key parameters for each user query ($Q$). The workflow is complete if the calculation request is successfully resolved in a \emph{zero-shot} scenario. Otherwise, the request proceeds to the \texttt{AEH} subsystem, which can be a successful case. Given an effective recommendation system, we can decompose $S$ as in Eq. \ref{eq:with_rec}:
\begin{equation}
P(S)=P(ZS)+\bigl(1-P(ZS)\bigr)\,P\!\left(AEH\mid\neg ZS\right).
\label{eq:with_rec}
\end{equation}

To estimate the system’s performance in the absence of $Rec$, we introduce the scaling factors $\alpha$ and $\beta$, which quantify how much $Rec$ multiplies the \emph{zero-shot} and \texttt{AEH} success probabilities, respectively, assuming $\alpha, \beta \geq 1$. Using Eq. \ref{eq:with_rec} the hypothetical $Q$-only success probabilities are then,
\begin{equation}
P(ZS\mid Q\text{-only})=\frac{0.1424}{\alpha},\qquad
P\!\left(\texttt{AEH}\mid\neg ZS,\,Q\text{-only}\right)=\frac{0.7628}{\beta},
\label{eq:without_alpha_beta}
\end{equation}
so that
\begin{equation}
P(S\mid Q\text{-only})
  =\frac{0.1424}{\alpha}+\frac{0.7628}{\beta}-\frac{0.1086}{\alpha\beta}.
\label{eq:without_tot}
\end{equation}
Setting \(\alpha=\beta=\gamma\) gives
\begin{equation}
P(S\mid Q\text{-only})=\frac{0.9052}{\gamma}-\frac{0.1086}{\gamma^{2}}.
\label{eq:without_tot_simp}
\end{equation}

This dependency of the success probability on the effectiveness of the recommendation system can be considered with a few representative examples: In the limiting case where the recommendation system has no effect ($\gamma = 1$), the success probability is \num{0.7966}, which is the same as the overall success probability with the recommendation system. A case where the recommendation system has an effectiveness reduction of \SI{50}{\percent} (i.e., $\gamma = 1.50$), the success rate without the recommendation system drops to \num{0.56}. In the case of double effectiveness ($\gamma =2$), the success rate further declines to \num{0.43}. The sensitivity of the success rate concerning variations in the recommendation system's effectiveness is shown in the following Eq. \eqref{eq:derivative}:
\begin{equation}
\frac{d}{d\gamma}P(S \mid Q\text{-only}) = -\frac{0.9052}{\gamma^2} + \frac{0.2172}{\gamma^3}, \quad \text{for } \gamma > 1.
\label{eq:derivative}
\end{equation}

This derivative indicates that the reduction in success rate (when the recommendation system is removed) is most sensitive when its effectiveness factor ($\gamma$) is close to $1$. These results imply that the recommendation system significantly boosts system performance. As $\gamma$ increases (implying that the recommendation system is even slightly effective), the success probability in the $Q$-only regime diminishes sharply. The derivative analysis confirms that the decrease in success probability is steepest when $\gamma$ is near 1. Small enhancements due to the recommendation system can lead to substantial differences in overall performance.

\section*{Conclusion}

Our study demonstrates that the GENIUS framework substantially improves productivity in DFT calculation, streamlining the entire process from typing a query to running the simulations. By marrying a domain-specific knowledge graph with a tiered stack of large language models and an intelligent-automated error-handling loop, the framework converts free-form user requests into validated Quantum ESPRESSO inputs, achieving successful execution on first attempt approximately \SI{80}{\percent} of the time. When the initial attempt fails, the agentic loop repairs $>\SI{76}{\percent}$ of crashes, and the attempt-wise success curve follows a fast exponential decay toward a stable \SI{7}{\percent} baseline, evidence that most recoverable errors are neutralized in the earliest retries. Three architectural choices underpin this reliability and efficiency. \textbf{Smart knowledge graph}:  Encapsulating \num{247} parameters and \num{330} dependency edges, the KG supplies the LLMs with grounded, constraint-aware facts, sharply reducing hallucinations and ensuring syntactic and physical consistency. \textbf{Model hierarchy}: Lightweight models tackle the bulk of queries, while larger models are invoked only when the workflow stalls, cutting inference cost and latency without sacrificing accuracy. \textbf{Finite-state error recovery}. A transparent finite-state machine monitors every run, restarts from a clean template after three failed fixes, and escalates only when the evidence justifies the expense. Together, these elements close the \emph{know–do gap} that separates mature electronic-structure codes from routine, accessible usage. Experimentalists no longer detour into arcane input syntax, and computational specialists can redirect effort from boilerplate scripting to scientific exploration. In the context of ICME, GENIUS removes a critical implementation barrier, allowing predictive simulation to flow unimpeded into design loops and high-throughput campaigns. By automating protocol generation, validation, and repair, GENIUS democratizes access to advanced simulation tools for groups lacking deep computational expertise, enabling wider participation in materials discovery. Its cost-aware orchestration makes large-scale screening feasible on moderate budgets, and its transparent logs assure reproducibility, a prerequisite for FAIR data practice. The present KG covers only \texttt{pw.x}; extending it to other Quantum ESPRESSO modules and codes beyond QE will broaden GENIUS’s reach. Community-driven contributions could add more simulation codes, refine edge conditions, and enrich metadata that remain manually curated. Finally, incorporating physics-informed validators (e.g., symmetry checks, charge counting) and adaptive hyperparameter tuning promises an additional jump in first-shot accuracy. GENIUS shows that the long-standing technical drag on computational materials science can be lifted when factual domain knowledge, strategic model selection, and disciplined workflow control are fused into a single agentic system.

\newpage
\section*{Declaration of Competing Interest}

The authors declare that they have no known competing financial interests or personal relationships that could have appeared to influence the work reported in this paper.

\section*{Data Availability Statement} 

The authors confirm that the data supporting the study's findings are available in the article GitHub repository \url{https://github.com/KIT-Workflows/agentic-workflow-framework}, and upon request.

\section*{Acknowledgements}
We acknowledge support by the KIT Publication Fund of the Karlsruhe Institute of Technology. The authors are also thankful for financial support from the National Council for Scientific and Technological Development (CNPq, grant numbers 444431/2024-1, and 444069/2024-0). W.W. and C.R.C.R. thank the German Federal Ministry of Education and Research (BMBF) for financial support of the project Innovation-Platform MaterialDigital (\url{www.materialdigital.de}) through project funding FKZ number 13XP5094A.

\newpage

\bibliography{jshort_apr2023,boxref_nat}

\end{document}